\documentclass[letterpaper,twocolumn]{article}
\usepackage{aaai19}
\usepackage{times}
\usepackage{helvet}
\usepackage{courier}
\usepackage{ulem}

\nocopyright

\usepackage{graphicx}
\usepackage{enumitem}
\usepackage{amsmath}
\usepackage{color}

\usepackage{bigstrut}
\usepackage{array}

\usepackage{natbib}

\renewcommand\cite{\citep}
\newcommand{\namecite}{\citet}

\usepackage{url}

\newcommand\aristobert{AristoBERT}
\newcommand\roberta{RoBERTa}
\newcommand\aristoroberta{AristoRoBERTa}

\newcommand{\eat}[1]{}
\mathchardef\mhyphen="2D
\newenvironment{ite}{                     
     \parskip 0cm \begin{itemize} \parskip 0cm \parsep 0cm \itemsep 0cm \topsep 0cm}{
        \end{itemize}} 

\usepackage{quoting}

\title{From `F' to `A' on the N.Y.~Regents Science Exams: \\ An Overview of the Aristo Project}

\author{Peter Clark, Oren Etzioni,  Daniel Khashabi, Tushar Khot, Bhavana Dalvi Mishra, \vspace{1mm} \\
{\bf \Large Kyle Richardson, Ashish Sabharwal, Carissa Schoenick, Oyvind Tafjord, Niket Tandon,} \vspace{1mm}\\
{\bf \Large Sumithra Bhakthavatsalam, Dirk Groeneveld, Michal Guerquin, Michael Schmitz} \vspace{1mm} \\
\ \\
Allen Institute for Artificial Intelligence, Seattle, WA, U.S.A.}

\date{}

\begin{document}
\maketitle

\begin{abstract}

  AI has achieved remarkable mastery over games such as Chess, Go, and Poker, and even {\it Jeopardy!}, but the rich variety of standardized exams has remained a landmark challenge.  Even as recently as 2016, the best AI system could achieve merely 59.3\% on an 8th Grade science exam \cite{Schoenick2016MovingBT}.

This article reports success on the Grade 8 New York Regents Science Exam,
where for the first time a system scores more than 90\% on the exam's non-diagram, multiple choice (NDMC)
questions. In addition, our Aristo system, building upon the success of recent language models,
exceeded 83\% on the corresponding Grade 12 Science Exam NDMC questions. The results, on unseen test questions, are robust across different test years and different variations of this kind of test. They demonstrate that modern Natural Language Processing (NLP) methods can result in mastery on this task. While not a full solution to general question-answering
(the questions are limited to 8th Grade multiple-choice science) it represents a significant milestone for the field.

\end{abstract}


In 2014, Project Aristo was launched with the goal of reliably answering grade-school science
questions, a stepping stone in the quest for systems that understood and could reason about science.
The Aristo goal was highly ambitious, with the initial system scoring well below
50\% even on 4th Grade multiple choice tests. With a glance at the questions,
it is easy to see why: the questions are hard. For example, consider the following
8th Grade question:
\begin{quote}\fbox{\parbox{0.9\columnwidth}{
{\it How are the particles in a block of iron affected when the block is melted?}\\
{\it (A) The particles gain mass.}\\
{\it (B) The particles contain less energy.}\\
{\it (C) The particles move more rapidly.} {\bf [correct]} \\
{\it (D) The particles increase in volume.}}}
\end{quote}
This question is challenging as it requires both scientific knowledge
(particles move faster at higher temperatures) and common sense knowledge (melting involves
raising temperature), and the ability to combine this information together appropriately.

\begin{figure}
\begin{center}
{\includegraphics[width=\columnwidth]{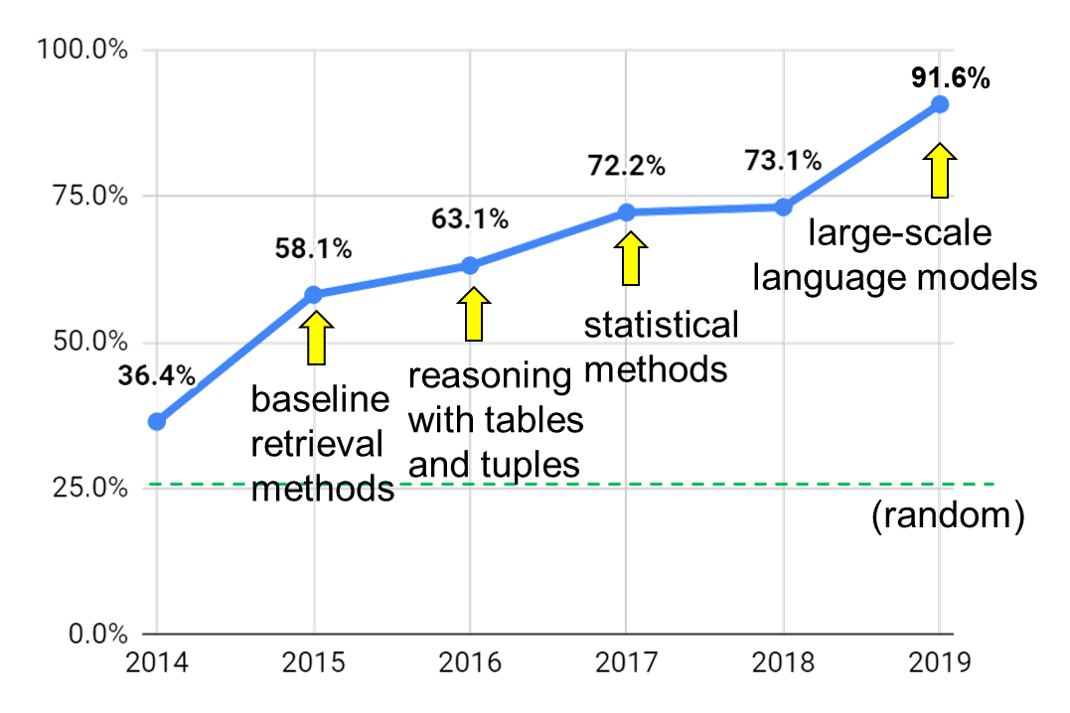}}
\end{center}
\caption{Progress over time of Aristo's scores on Regents 8th Grade Science (non-diagram, multiple choice questions, held-out test set).}
\label{progress}
\end{figure}

Now, six years later, we are able to report that Aristo recently surpassed  90\% on multiple
choice questions from the Grade 8 New York Regents Science Exam, a major milestone
and a reflection on the tremendous progress of the NLP community as a whole.
In this article, we review why this is significant, how Aristo was able to achieve this score,
and where the system still makes mistakes. We also explore what kinds of reasoning Aristo appears to be capable of 
doing, and what work still needs to be done to achieve the broader goals of the project.

\eat{
  ORIGINAL SUBMITTED TEXT
  Why is this an important achievement? First, standardized tests, in particular science exams,
are a rare example of a good benchmark for AI. Machine intelligence is most appropriately
viewed as a diverse collection of capabilities rather than a single skill. As a result, a good benchmark
should test a variety of capabilities while also being clearly measurable,
understandable, accessible, non-gameable, and sufficiently motivating. Standardized tests,
while not a full test of machine intelligence, largely meet these requirements and explore several capabilities strongly associated with intelligence including language
understanding, reasoning, and the use of common sense knowledge \cite{aimagazine2016}.
}

Why is this an important achievement? First, passing standardized tests has
been a challenging AI benchmark for many years \cite{bringsjord2003artificial,brachman2005selected,strickland2013can}.
A good benchmark should test a variety of capabilities while also being clearly measurable,
understandable, accessible, non-gameable, and sufficiently motivating. Standardized tests,
while not a full test of machine intelligence, meet many of these practical
requirements \cite{clark2016my}. They also appear to require several
capabilities associated with intelligence, including language understanding,
reasoning, and commonsense - although the extent to which such 
skills are needed is controversial \cite{Davis2014TheLO}. We
explore this in more detail this article.

\eat{
Why is this an important achievement? First, standardized tests, in particular
science exams, meet many of the criteria of a good AI benchmark. In particular, 
a good benchmark should test a variety of capabilities while also being clearly measurable,
understandable, accessible, non-gameable, and sufficiently motivating. Standardized tests,
while not a full test of machine intelligence, meet many of these practical
requirements \cite{aimagazine2016}. They also appear to require several
capabilities associated with intelligence, including language understanding,
reasoning, and commonsense - although the extent to which such 
skills are needed remains controversial \cite{Davis2014TheLO}. We
explore this in more detail this article.}

\eat{
Why is this an important achievement? First, standardized tests, in particular science exams,
meet many of the criteria of a good AI benchmark - although with caveats, as we discuss shortly.
Machine intelligence is most appropriately viewed as a diverse collection of capabilities rather
than a single skill. As a result, a good benchmark should test a variety of capabilities while also being clearly measurable,
understandable, accessible, non-gameable, and sufficiently motivating. Standardized tests,
while not a full test of machine intelligence, meet many of these requirements \cite{aimagazine2016}.
}

Second, although NLP has made dramatic advances in recent years with
the advent of large-scale language models such as ELMo \cite{elmo}, BERT \cite{bert},
and \roberta{} \cite{roberta}, many of the demonstrated successes
have been on internal yardsticks generated by the AI/NLP community itself,
such as SQuAD \cite{Rajpurkar2016SQuAD10}, GLUE \cite{wang2018glue}, 
and TriviaQA \cite{JoshiTriviaQA2017}. In contrast, the 8th Grade science exams are an external,
independently-generated benchmark where we can compare machine performance with human performance.
Aristo thus serves as a ``poster child" for the remarkable and rapid advances achieved in
NLP, applied to an easily accessible task.

Finally, Aristo makes steps towards the AI Grand Challenge of a system that can read a textbook chapter
and answer the questions at the end of the chapter. This broader challenge dates back to the 1970s,
and was reinvigorated in Raj Reddy's 1988 AAAI Presidential Address and subsequent
writing \cite{reddy1988foundations,Reddy2003ThreeOP}. 
However, progress on this challenge has a checkered history.
Early attempts side-stepped the natural language understanding (NLU) task, in the belief
that the main challenge lay in problem solving, e.g., \cite{Larkin1980ModelsOC}.
In recent years there has been substantial progress in systems that can find
factual answers in text, starting with IBM's Watson system \cite{watson}, and
now with high-performing neural systems that can answer short questions
provided they are given a text that contains the answer \citep[e.g.,][]{Seo2016BidirectionalAF,Wang2018MultigranularityHA}.
Aristo continues along this trajectory, but aims to also answer
questions where the answer may not be written down explicitly.
While not a full solution to the textbook grand challenge, Aristo
is a further step along this path.

At the same time, care is needed in interpreting Aristo's results. In particular,
we make no claims that Aristo is answering questions in the way a person would
(and is likely using different methods). 
Exams are designed with human reasoning in mind, to test certain human knowledge
and reasoning skills. But if the computer is answering questions in a
different way, to what extent does it possess such skills? \cite{Davis2014TheLO}.
To explore this, we examine the causes of some of Aristo's failures, and 
test whether Aristo has some of the semantic skills that appear
necessary for good performance. We find evidence of several types of such
systematic behavior, albeit not perfect, suggesting {\it some} form of
reasoning is occurring. Although still quite distant from human problem-solving,
these emergent semantic skills are likely a key contributor to Aristo's scores
reaching the 90\% range.

As a brief history, the metric progress of the Aristo system on the Regents 8th Grade exams (non-diagram,
multiple choice part, for a hidden, held-out test set) is shown in Figure~\ref{progress}.
The figure shows the variety of techniques attempted, and mirrors the rapidly
changing trajectory of the Natural Language Processing (NLP) field in general.
Early work was dominated by information retrieval, statistical, and automated
rule extraction and reasoning methods
\cite{clark2014:akbc,clark2016combining,khashabi2016tableilp,Khot2017AnsweringCQ,Khashabi2018QuestionAA}.
Later work has harnessed state-of-the-art tools for large-scale language modeling
and deep learning \cite{trivedi2019repurposing,tandon2018reasoning}, which
have come to dominate the performance of the overall system and reflects
the stunning progress of the field of NLP as a whole.

Finally, it is particularly fitting to report this result in the
AI Magazine, as it is another step in the decades-long
quest to fulfil the late Paul Allen's dream of a ``Digital Aristotle'',
an ``easy-to-use, all-encompassing knowledge storehouse...to advance the field
of AI.'' \cite{allen2012idea}, a dream also set out in the AI Magazine
in the Winter 2004 edition \cite{friedland2004project}. Aristo's
success reflects how much progress the field of NLP and AI as a whole
has made in the intervening years.

\begin{figure}
\begin{center}
{\includegraphics[width=\columnwidth]{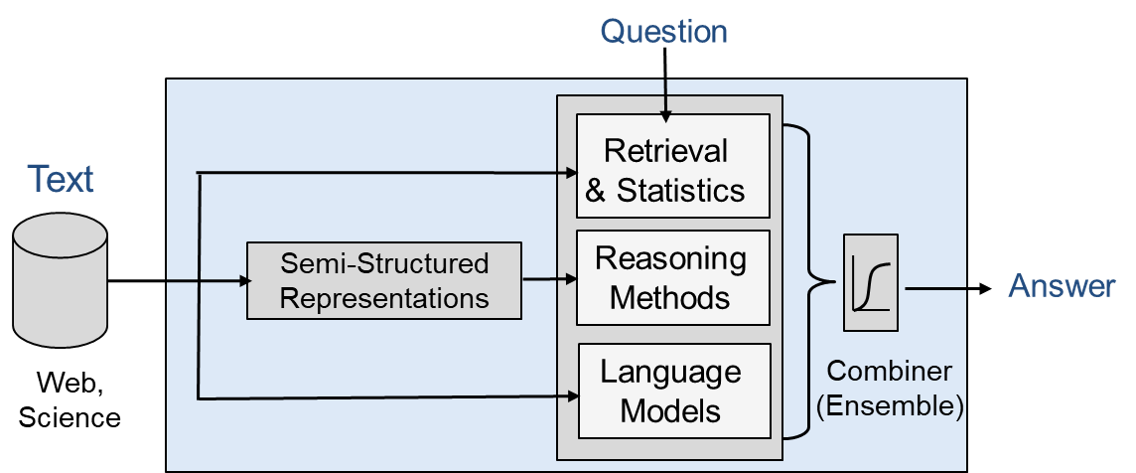}}
\end{center}
\caption{A simplified picture of Aristo's architecture. Aristo's eight solvers can be loosely grouped into
  statistical, reasoning, and language model-based approaches. \label{architecture}}
\end{figure}

\section{The Aristo System}

Aristo comprises of eight solvers, each of which attempts to independently answer a multiple choice question.
Its suite of solvers has changed over the years, with new solvers being added and redundant
solvers being dropped to maintain a simple architecture. (Earlier solvers include use of
Markov Logic Networks \cite{Khot2015ExploringML}, reasoning over tables \cite{khashabi2016tableilp},
and other neural approaches, now superceded by the language models.) As illustrated in Figure~\ref{architecture}, they can be loosely grouped into:
\begin{enumerate}
\item Statistical and information retrieval methods
\item Reasoning methods
\item Large-scale language model methods
\end{enumerate}
We now briefly describe these solvers, with pointers to further information.
Over the life of the project, the relative importance of the methods has shifted towards large-scale language methods,
which now dominate the overall performance of the system.

\eat{
  \begin{figure}
\begin{center}
{\includegraphics[width=0.8\columnwidth]{multee.png}}
\end{center}
  \caption{Multee retrieves potentially relevant sentences, then for each answer option
    in turn, assesses the degree to which each sentence entails that answer.
    A multi-layered aggregator then combines this (weighted) evidence from each sentence.
    In this case, the strongest overall support is found for option ``(C) table salt'', so it is selected.}
\label{multee}
\end{figure}
}

\subsection{Information Retrieval and Statistics}

The {\bf IR solver} searches to see if the question along with an answer option is explicitly stated in the corpus.
To do this, for each answer option $a_i$, it sends $q$ + $a_i$
as a query to a search engine, and returns the search engine's score for the top retrieved sentence
$s$. This is repeated for all options $a_i$ to score
them all, and the option with the highest score selected \cite{clark2016combining}.

The {\bf PMI solver} uses pointwise mutual information \cite{church1989} to measure the associations between parts of $q$ and parts of $a_i$. PMI for two n-grams $x$ and $y$ is defined as
$\mathrm{PMI}(x,y) = \log \frac{p(x,y)}{p(x) p(y)}$.
The larger the PMI, the stronger the association between $x$ and $y$. The solver extracts unigrams, bigrams, trigrams, and skip-bigrams,
and outputs the answer with the largest average PMI, calculated over all pairs of question and answer option n-grams \cite{clark2016combining}.

Finally, {\bf ACME} (Abstract-Concrete Mapping Engine) searches for a cohesive link between
a question $q$ and candidate answer $a_{i}$ using a large knowledge base of 
{\it vector spaces} that relate words in language to a set of 5000 scientific terms enumerated
in a {\it term bank}. The key insight in ACME is that we can better assess lexical cohesion of a question and
answer by pivoting through scientific terminology, rather than by simple co-occurence 
frequencies of question and answer words \cite{turney2017leveraging}.

\subsection{Reasoning Methods}

\begin{figure}
\begin{center}
{\includegraphics[width=\columnwidth]{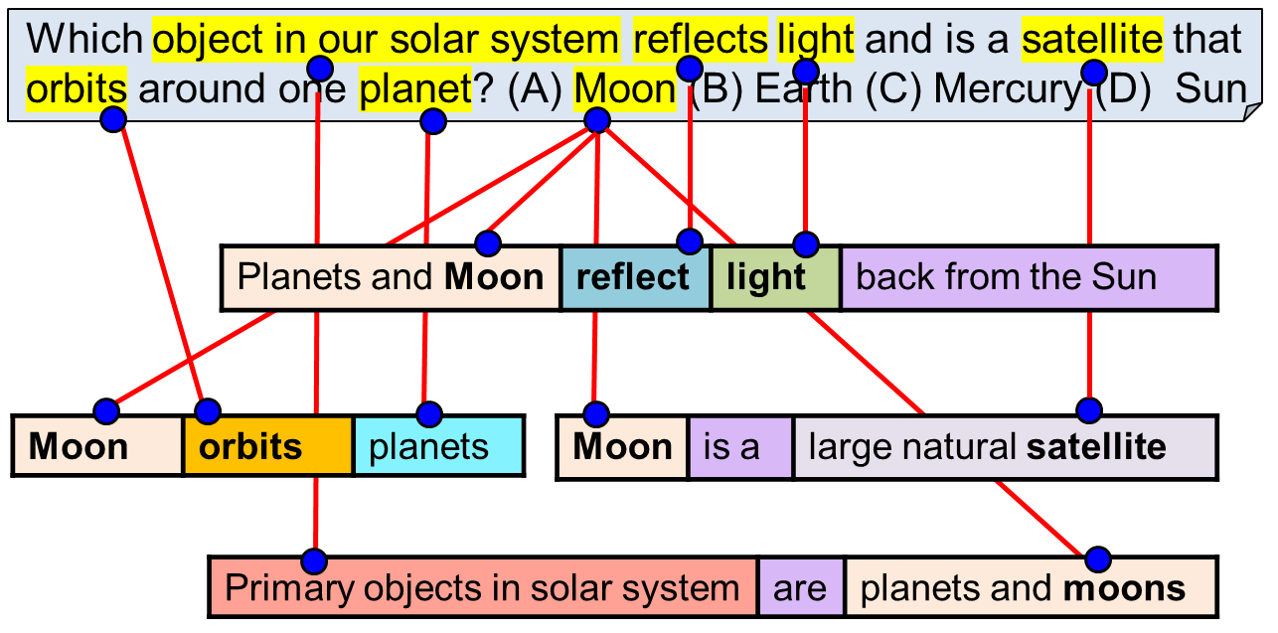}}
\end{center}
\caption{The TupleInference Solver retrieves tuples relevant to the question, and constructs a
  support graph for each answer option (Here, the support graph for the choice (A) is shown).}
\label{fig:suppGraph}
\end{figure}

The {\bf TupleInference solver} uses semi-structured knowledge in the form of {\it tuples},
extracted via Open Information Extraction (Open IE)~\cite{Banko2007OpenIE}.
TupleInference treats the reasoning task as searching for a graph that best connects the terms in the question with an answer choice via the knowledge; see Figure~\ref{fig:suppGraph} for a simple illustrative example. To find the best support graph for each answer option, we define the task as an optimization problem,
and use Integer Linear Programming (ILP) to solve it. 
The answer choice with the highest scoring graph is then selected \cite{Khot2017AnsweringCQ}.

{\bf Multee} \cite{trivedi2019repurposing} is a solver that repurposes existing {\it textual entailment} tools for question answering.
Textual entailment (TE) is the task of assessing if one text implies another,
and there are several high-performing TE systems now available.
Multee learns to combine their decisions, so it can determine how strongly a set of retrieved texts entails a particular question + answer option
\cite{trivedi2019repurposing}.

The {\bf QR (qualitative reasoning) solver} is designed to answer questions about qualitative influence, i.e., how more/less of one quantity affects another. 
Unlike the other solvers in Aristo, it is a specialist solver that only fires for a small subset of questions that ask about qualitative change.
The solver uses a knowledge base of 50,000 (textual) statements about qualitative influence, e.g., ``A sunscreen with a higher SPF protects the skin longer.''.
It has then been trained to reason using the BERT language model \cite{bert}, using a similar approach to that described below \cite{quartz}.

\subsection{Large-Scale Language Models}

\begin{table}
  \centering
  {\small
  \begin{tabular}{|l|lll|l|} \hline
   & \multicolumn{3}{|c|}{\bf Partition} & \\
 {\bf Dataset} & {\bf Train} & {\bf Dev} & {\bf Test} & {\bf Total} \\ \hline
Regents 4th   &              127 &    20 &    109  &  256 \\
Regents 8th    &             125 &    25 &    119 & 269 \\
Regents 12th      &         665 &  282 &    632 & 1579\\
ARC-Easy         & 2251 &  570 &  2376 & 5197 \\
ARC-Challenge     & 1119  &  299 &  1172 & 2590 \\ \hline
Totals$^{\dag}$ &  4035 & 1151 & 4180 & 9366 \\ \hline
  \end{tabular} \\
                {\small  $^{\dag}$ARC (Easy+Challenge) includes Regents 4th \& 8th as a subset}
                }
\vspace{-3mm}
  \caption{Dataset partition sizes (number of questions).   \label{dataset-sizes}}
\end{table}

The field of NLP has advanced substantially with the advent of large-scale language
models such as ELMo \cite{elmo},
BERT \cite{bert}, and \roberta{} \cite{roberta}.
The {\bf \aristobert}~solver applies BERT
to multiple choice questions by treating the task as classification:
Given a question $q$ with answer options $a_{i}$ and optional background 
knowledge $K_{i}$, we provide it to BERT as:
\begin{quote}
  {\it [CLS] $K_i$ [SEP] $q$ [SEP] $a_{i}$ [SEP]}
\end{quote}
for each option $a_i$.
The [CLS] output token is projected to a single logit and fed
through a softmax layer across answer options, trained using cross entropy loss,
and the highest scoring option selected.

\aristobert~uses three methods to apply BERT more effectively.
First, we retrieve and supply background knowledge $K_i$ along with the question when using BERT,
as described above. This provides the potential for BERT to ``read'' that background knowledge and
apply it to the question, although the exact nature of how it uses background
knowledge is more complex and less interpretable. Second, following
\cite{sun2018improving}, we
fine-tune BERT using a curriculum of several datasets, starting with 
RACE (a general reading comprehension dataset that is not science related \cite{race}),
followed by a collection of science training sets: OpenbookQA
\cite{Mihaylov2018CanAS}, ARC \cite{Clark2018ThinkYH}, and
Regents questions (training partition). Finally, 
we repeat this for three variants of BERT (cased, uncased, cased whole-word), and ensemble the predictions.

Finally, the {\bf \aristoroberta}~solver does the same with RoBERTa \cite{roberta}, a high-performing and optimized derivative of BERT trained on significantly more text.
In AristoRoBERTa, we simply replace the BERT model in \aristobert{} with \roberta{}, repeating similar fine-tuning steps. We ensemble two versions together, namely with and without the first fine-tuning step using RACE.

\begin{figure*}
\begin{center}
{\includegraphics[width=\textwidth]{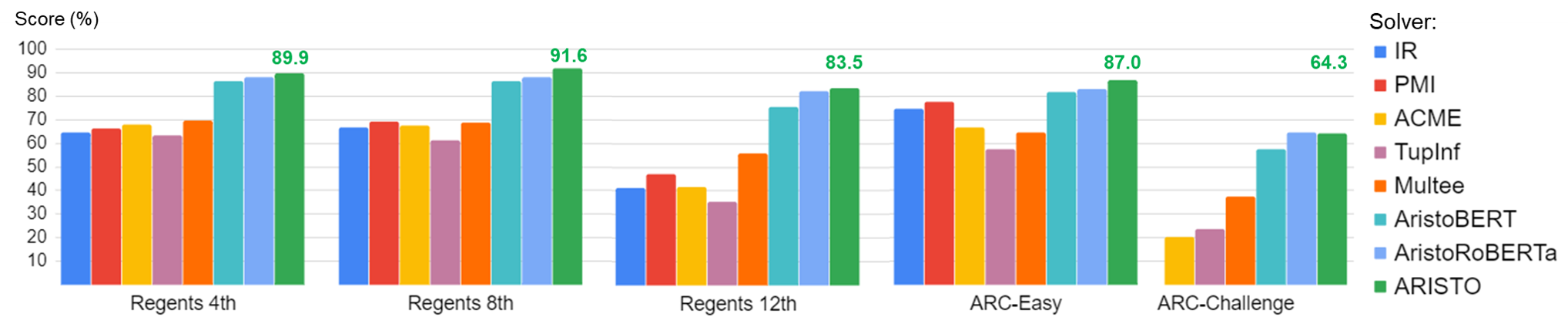}}
\end{center}
 \caption{The results of each of the Aristo solvers, as well as the overall Aristo system, on each of the test sets.  Most notably, Aristo achieves 91.6\% accuracy in 8th Grade, and exceeds 83\% in 12th Grade. Note that Aristo is a single system, run unchanged on each dataset (not re-tuned for each dataset).}
\label{results}
\end{figure*}

\section{Experiments and Results}

We apply Aristo to the non-diagram, multiple choice (NDMC) questions in the science exams.
Although questions with diagrams are common,\footnote{
  Ratios of non-diagram multiple choice (NDMC), with-diagram multiple-choice (DMC), non-diagram direct answer (NDDA),
  and with-diagram direct answer (DDA) questions are approximately 45/25/5/25 for Regents 4th Grade, 25/25/5/45 for 8th Grade,
  and 40/25/15/20 for 12th grade.}
they are outside of our focus on language and reasoning.
(For illustrative work on science diagrams, see \cite{Krishnamurthy2016SemanticPT}).
We also omit questions that require a direct answer for two reasons. First, after removing questions with diagrams,
they are rare: For example, of the 482 direct answer questions over 13 years of Regents 8th Grade Science exams, only 38 ($<$8\%)
do {\it not} involve a diagram. Second, they are complex, often requiring explanation and synthesis.
Both diagram and direct-answer questions are natural topics for future work.

We evaluate Aristo using the New York Regents Science exam questions\footnote{See https://www.nysedregents.org/ for the original exams.},
and also the ARC dataset, a larger corpus of science questions drawn from public resources across the country \cite{Clark2018ThinkYH}.
The Regents exams are only produced for 4th, 8th, and 12th Grade students (corresponding to the end of elementary, middle, and
high-school respectively), while the ARC questions span Grades 3 to 9.
All questions are posed exactly as written, with no rewording or rephrasing. The entire dataset is partitioned
into train/dev/test parts (Table~\ref{dataset-sizes}), and for the Regents questions we ensure that each exam is either completely in train, dev, or test but
not split between them. (The non-Regents ARC questions do not have exam groupings.) 
All but 39 of the 9366 questions are 4-way multiple choice, the remaining 39 ($<$0.5\%) being 3- or 5-way.
A random score over the entire dataset is 25.02\%.

\subsection{Results}

The results are summarized in Figure~\ref{results}, showing the performance of the solvers individually, and their combination in
the full Aristo system. Note that Aristo is a single system run on the five datasets (not retuned for each dataset in turn).

Most notably, Aristo's scores on the Regents Exams far
exceed earlier performances \cite[e.g.,][]{Schoenick2016MovingBT,clark2016combining}, and represent a
new high-point on science questions.

In addition, the results show the dramatic impact of new
language modeling technology, embodied in AristoBERT and AristoRoBERTa, the scores for these two
solvers dominating the performance of the overall system. Even on the ARC-Challenge questions,
containing a wide variety of difficult questions, the language modeling based solvers dominate.
The general increasing trend of solver scores from left to right for each test set loosely
reflects the progression of the NLP field over the six years of the project.

To further check that we have not overfit to our data, we also ran Aristo on the most recent years
of the Regents Exams (4th and 8th Grade), years 2017-19, that were unavailable at the start of
the project and were not part of our datasets. We find similar scores (average 92.8\% for the
three Fourth Grade exams, 93.3\% for the Eighth Grade), suggesting the system is not
overfit.

On a combination of exam scores and lab work (weighted approximately 60:40), the NY State Education Department considers an overall score of 65\% as
``Meeting the Standards'', and over 85\% as ``Meeting the Standards with Distinction''\footnote{
  https://www.nysedregents.org/grade8/science/618/home.html}.
As a somewhat loose comparison, if this rubric applies equally to the NDMC subset we have studied, this would mean Aristo has
met the standard with distinction in 8th Grade NDMC Science (although clearly full science requires substantially more).

\subsection{Answer Only Performance}

\begin{figure}
\begin{center}
{\includegraphics[width=\columnwidth]{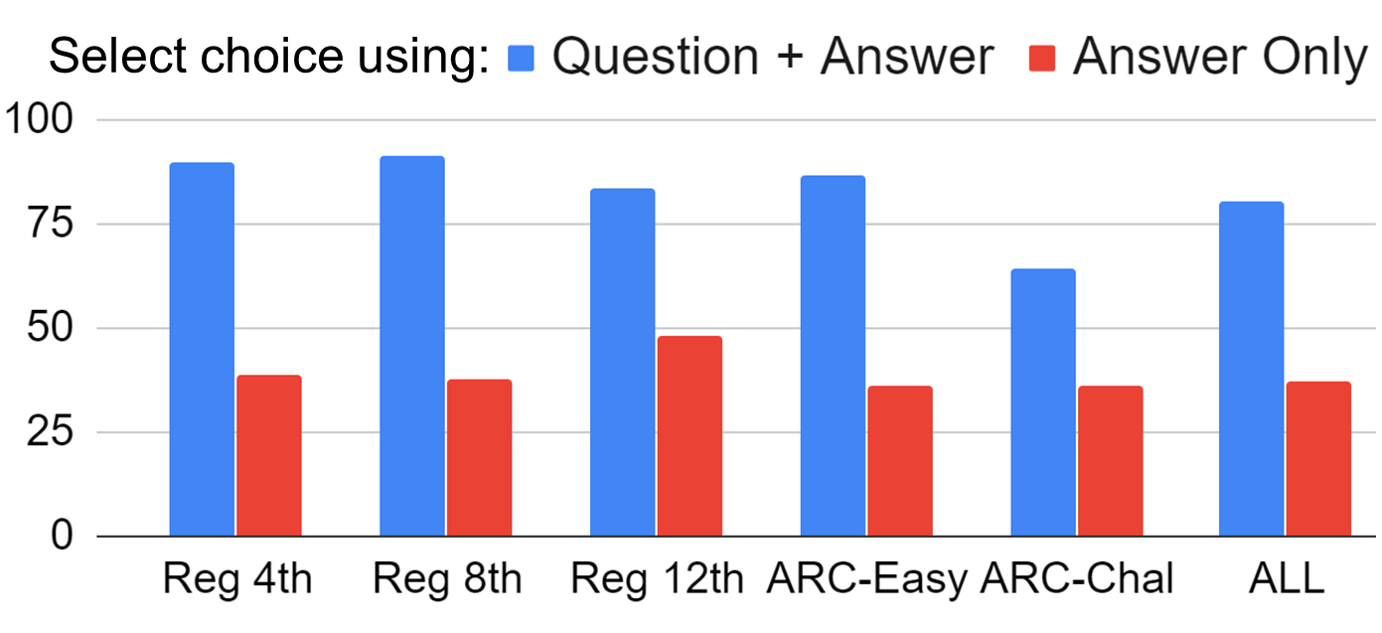}}
\end{center}
\caption{Scores when looking at the answer options only, compared with using the full questions.
  The (desirably) low scores/large drops indicate it is hard to guess the answer without reading the question.
  \label{answer-only-table}}
\end{figure}

Several authors have observed that for some multiple choice datasets, systems can still perform well
even when ignoring the question body and looking only at the answer options \cite{gururangan2018annotation,poliak2018hypothesis}.
This surprising result is particularly true for crowdsourced datasets, where workers may
use stock words or phrases (e.g., ``not'') in incorrect answer options that gives them away.
To measure this phenomenon on our datasets, we trained and tested a new AristoRoBERTa model
giving it only the answer options, i.e., no question body nor retrieved knowledge. (Without
retraining the model scores slightly less, 35\% overall vs. 38\% with retraining).
The results (test set) are shown in Figure~\ref{answer-only-table}, indicating that it
is hard to select the right answer without reading the question.
(Scores are slightly higher for 12th Grade answer-only, possibly because
the average answer length is longer, hence more potential for
hidden patterns inside that hint at correctness/incorrectness).

\subsection{Adversarial Answer Options}

\begin{figure}
\begin{center}
{\includegraphics[width=\columnwidth]{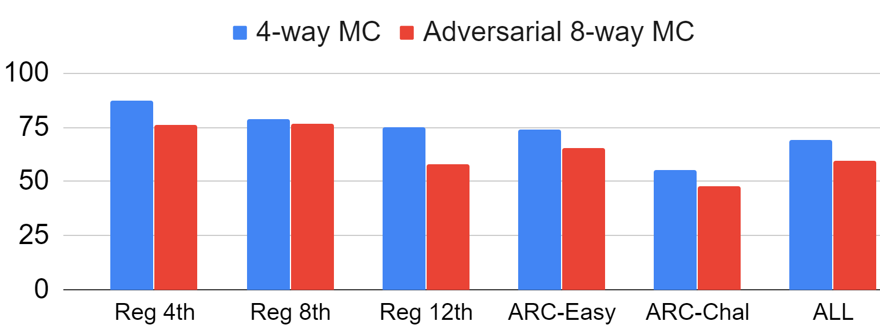}}
\end{center}
\caption{Aristo's scores drop a small amount (average 10\%) when tested
  on adversarially generated 8-way MC questions.}
  \label{adversarial}
\end{figure}

What if we add extra incorrect answer options to the questions? If a system has mastery of the material,
we would expect its score to be relatively unaffected by such modifications.
We can make this more challenging by doing this {\it adversarially}: try many
different incorrect options until the system is fooled. If we do this,
turning a 4-way MC question into 8-way with options chosen to fool Aristo, 
then retrain on this new dataset, we do observe an effect: the scores
drop, although the drop is small ($\approx$10\%), see Figure~\ref{adversarial}.
\eat{
For example, while the solver gets the right answer to the following question:
\begin{quote}\fbox{\parbox{0.9\columnwidth}{
  {\it The condition of the air outdoors at a certain time of day is known as
    (A) friction (B) light (C) force (D) weather {\bf [selected, correct]}}}}
\end{quote}
it fails for the 8-way variant:
\begin{quote}\fbox{\parbox{0.9\columnwidth}{
{\it The condition of the air outdoors at a certain time of day is known as
  (A) friction (B) light (C) force (D) weather {\bf [correct]} (Q) joule (R) gradient {\bf [selected]} (S) trench (T) add heat}}}
\end{quote}
}
This indicates that while Aristo performs well, it still has some blind spots that can be
artificially uncovered through adversarial methods such as this.

\section{Analysis \label{failures}}

\begin{figure}
\begin{center}
{\includegraphics[width=0.8\columnwidth]{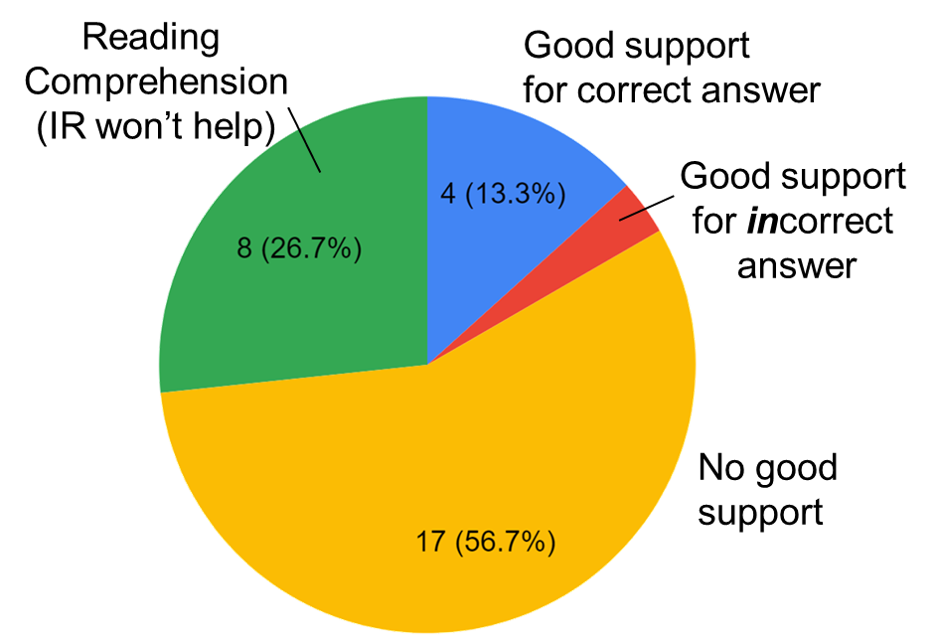}}
\end{center}
\caption{Case Study: Causes of error for 30 questions that Aristo answered incorrectly. \label{failure-pie-chart}}
\end{figure}

Despite the high scores, Aristo still makes occasional mistakes. Because
Aristo retrieves and then ``reads'' corpus sentences to answer
a question, we can inspect the retrieved knowledge when Aristo
fails, and gain some insight as to where and why it makes errors.
Did Aristo retrieve the ``right'' knowledge, but then choose the wrong
answer? Or was the failure due (in part) to the retrieval step itself?
We manually analyzed 30 random failures (of 248) in
the entire dev set (Regents + ARC, 1151 dev set questions total), and found four main
categories of failures, illustrated in Figure~\ref{failure-pie-chart},
that we now summarize.
As the language model solvers have highest weight in
Aristo, we conduct this analysis for failures by AristoRoBERTa, but note
these very frequently ($\sim$90\% of the time) equate to overall
Aristo failures, and that when AristoRoBERTa fails, 
most (on average, 76\%) of the other solvers also fail also. We
did not discern any systematic patterns within these.


\subsubsection{Good Support for Correct Answer (13\%)}

Surprisingly, only 4 of the 30 failures were cases where the
retrieved knowledge supported the right answer option, but Aristo
chose a wrong answer option. An example is:
\begin{quote}\fbox{\parbox{0.9\columnwidth}{
    {\it Which is the best unit to measure distances between Earth and 
      other solar systems in the universe? (A) miles (B) kilometers (C) light years {\bf [correct]} (D) astronomical units {\bf [selected]}}}}
  \end{quote}
  Here, although Aristo {\it did} retrieve good evidence for the correct answer (C), namely:
  \begin{quote}
    {\it Distances between Earth and the stars are often measured in terms of light-years.}
  \end{quote}
it still preferred the incorrect option (D) from the retrieved knowledge:
  \begin{quote}
    {\it In general, distances in the solar system are measured in astronomical units.}
  \end{quote}
  Here, Aristo has confused distinguishing distances {\it within} the Solar System vs. distances {\it between} solar systems.
  (A confusion that a human might easily make too). This illustrates where Aristo has mis-applied its
  retrieved knowledge. However, such cases appear to be rare (4/30). In other words,
  for the fast majority of questions, {\it if} suitable knowledge is retrieved, then Aristo 
  will answer correctly.

  \subsubsection{No Support for the Correct Answer (57\%)}

  The largest cause of failure was simply when none of the retrieved sentences
  provide evidence for the correct answer. In such situations, Aristo has little chance
  of answering correctly. For example:
\begin{quote}\fbox{\parbox{0.9\columnwidth}{
{\it Although they belong to the same family, an eagle and a pelican are different. What is one difference between them? (A) their preference for eating fish (B) their ability to fly (C) their method of reproduction {\bf [selected]} (D) their method of catching food {\bf [correct]}}}}
  \end{quote}
  As there are no corpus sentences comparing eagles and pelicans,
  Aristo retrieves a rather random collection of unhelpful facts.
  Instead, what is needed here is to realize this is a comparison question,
  then retrieve appropriate facts for pelicans and eagles individually,
  and them compare them, for example by using question decomposition
  methods \cite{Wolfson2020BreakID}.

\subsubsection{Reading Comprehension (27\%)}

In the exams, there are a few ``reading comprehension'' questions that primarily
require reasoning over the question content itself, rather than retrieving and applying
science knowledge. In such situations, retrieved knowledge is unlikely to be helpful.
8/30 failures fell into this category. One example is a question describing an experiment:
\begin{quote}\fbox{\parbox{0.9\columnwidth}{
  {\it A student wants to determine the effect of garlic on the growth of a fungus species. Several samples of fungus cultures are grown in the same amount of agar and light. Each sample is given a different amount of garlic. What is the independent variable in this investigation? (A) amount of agar (B) amount of light (C) amount of garlic {\bf [correct]} (D) amount of growth {\bf [selected]}}}}
  \end{quote}
  Here, the answer is unlikely to be written down in a corpus, as a novel scenario is being described.
  Rather, it requires understanding the scenario itself.

  A second example is a meta-question about sentiment:
\begin{quote}\fbox{\parbox{0.9\columnwidth}{
    {\it Which statement is an opinion? (A) Many plants are green.
      (B) Many plants are beautiful. {\bf [correct]} (C) Plants require sunlight. {\bf [selected]} (D) Plants can grow in different places.}}}
  \end{quote}
  Again, retrieval is unlikely to help here. Rather, the question asks for an analysis of the options themselves, something Aristo does not realize.

  \subsubsection{Good Support for an Incorrect Answer (3\%)}

  Occasionally a failure occurs due to retrieved knowledge supporting
  an incorrect answer, e.g., if the question is ambiguous, or the retrieved knowledge is wrong.
  The single failure in this category that we observed was:
\begin{quote}\fbox{\parbox{0.9\columnwidth}{
    {\it Which of these objects will most likely float in water? (A) glass marble (B) steel ball (C) hard rubber ball {\bf [selected]}
      (D) table tennis ball {\bf [correct]}}}}
  \end{quote}
Here, Aristo retrieved evidence for both (C) and (D), e.g., for (C), Aristo's retrieval included:
\begin{quote}
  {\it ``It had like a rubber ball in it, which would maybe float up…''}
\end{quote}
  here leading Aristo to select the wrong answer. Arguably, as this question is a comparative
  (``...which {\it most likely} floats?''), Aristo should have rejected this
  in favor of the correct answer ({\it table tennis ball}). However, as Aristo
  computes a confidence for each option independently, it is unable to directly
  make these cross-option comparisons.

  \subsubsection{Other}

  Finally we point to one other interesting failure:
\begin{quote}\fbox{\parbox{0.9\columnwidth}{
  {\it About how long does it take for the Moon to complete one revolution around Earth? (A) 7 days
    (B) 30 days {\bf [correct]} (C) 90 days (D) 365 days {\bf [selected]}}}}
\end{quote}  
In this case, many relevant sentences were retrieved, including:
\begin{ite}
\item {\it Because it takes the moon about {\bf 27.3 days} to complete one orbit around the Earth...}
\item {\it It takes {\bf 27.3 days} for the moon to complete one revolution around the earth.}
\item {\it The Moon completes one revolution around the Earth in {\bf 27.32166} days.}
\end{ite}
However, Aristo does not realize 27.3 is ``about'' 30, and hence answered the question incorrectly.

\section{A Score Card for Aristo's Semantic Skills \label{scorecard-section}}

From informal tests, 
Aristo appears to be doing more than simply matching a question + answer option to a retrieved sentence.
Rather, Aristo appears to recognize various linguistic and semantic phenomena, and respond appropriately.
For example, if we add negation (a ``not'') to the question, Aristo
almost always correctly changes its answer choice. Similarly if we replace ``increase'' with ``decrease''
in a question, Aristo will typically change its answer choice correctly,
suggesting it has some latent knowledge of qualitative direction.

To quantify such skills more systematically,
we performed five sets of tests on Aristo {\it without fine-tuning Aristo on those tests},
i.e., the tests are zero-shot. (From other experiments, we know that if we train Aristo on these
tests it can perform them almost perfectly, but our interest here is how Aristo
performs ``out of the box'' after training on the science exams).
Each test probes a different semantic phenomenon of interest, as we now describe.

\subsubsection{Negation}

How well does Aristo handle negation? As a (limited) test, we generated a synthetic negation dataset
(10k questions), where each question has a synthetic context (replacing the retrieved sentences),
plus a question about it, for example:
\begin{quote}\fbox{\parbox{0.9\columnwidth}{
  \underline{\bf Context:} \\
{\it  Alan is small. 
  Alan is tall. 
  Bob is big. 
  Bob is tall.  
  Charlie is big. 
  Charlie is tall. 
  David is small. 
  David is short.} 

  \underline{\bf Question:}   \\
  {\it Which of the following is {\bf not} tall? \\
    (A) Alan (B) Bob (C) Charlie (D) David {\bf [correct]}}}}
\end{quote}
We then test Aristo on this dataset {\it without fine-tuning on it}.
Remarkably, Aristo score 94\% on this dataset, suggesting
at least in this particular formulation, Aristo has an
understanding of ``not''.

\subsubsection{Conjunction}

We test conjunction in a similar way, with questions such as:
\begin{quote}\fbox{\parbox{0.9\columnwidth}{
  \underline{\bf Context:} \\
{\it Alan is red. 
Alan is big.   
Bob is blue. 
Bob is small. 
Charlie is blue. 
Charlie is big. 
David is red. 
David is small.}

  \underline{\bf Question:}   \\
  {\it Which of the following is big {\bf and} blue?
    (A) Alan (B) Bob (C) Charlie {\bf [correct]} (D) David}}}
\end{quote}
With questions containing two conjuncts (e.g., the one above), and
again {\it without} any fine-tuning on this data,
Aristo scores 98\%. If we increase the number of conjuncts
in the question to 3, 4, and 5, Aristo scores 95\%, 94\%, and 80\%
respectively. If we use five conjuncts {\it and} a negation,
for example:
\begin{quote}\fbox{\parbox{0.9\columnwidth}{
  \underline{\bf Context:} \\
{\it {\small
Alan is red. Alan is big. Alan is light. Alan is old. Alan is tall. Bob is red. Bob is small. Bob is heavy. Bob is old. Bob is tall. Charlie is blue. Charlie is big. Charlie is light. Charlie is old. Charlie is tall. David is red. David is small. David is heavy. David is young. David is tall. }}

\underline{\bf Question:}   \\
{\it {\small Which of the following is old {\bf and} red {\bf and} light {\bf and} big {\bf and not} short? (A) Alan (B) Bob (C) Charlie (D) David}}}}
\end{quote}
(the correct answer is left as an exercise for the reader) Aristo remarkably still scores 75\%.
Although not perfect, this indicates some form of systematic handling of conjunction plus negation is occurring.

\subsubsection{Polarity}

Polarity refers to Aristo's ability to correctly change its answer when a comparative
in the question is ``flipped''. For example, given:
\begin{quote}\fbox{\parbox{0.9\columnwidth}{
{\it Which human activity will likely have a negative effect on global stability?} \\
{\it (A) decreasing water pollution levels} \\
{\it (B) increasing world population growth.} {\bf [correct]}}}
\end{quote}
if we now switch ``negative'' to ``positive'', Aristo should switch its answer from (B) to (A). 
To score a point, Aristo must get {\it both} the original question
{\it and} the ``flipped'' question (with a changed answer) correct.
To measure this, we use an existing qualitative dataset containing
such pairs, called QuaRTz \cite{quartz}. As the QuaRTz questions are 2-way
multiple-choice, a random score for getting {\it both} right would be 25\%.
Remarkably, we find Aristo scores 67.1\% (again with no fine-tuning on QuaRTz),
suggesting Aristo has some knowledge about the polarity of comparatives.
Note that this test also requires Aristo to get the original question
right in the first place, thus the score reflects both knowledge
and polarity reasoning, a harder task than polarity alone.

\subsubsection{Factuality}

Event factuality refers to whether an event, mentioned in a textual context, did in fact
occur \cite{Rudinger2018NeuralMO}. For example:
\begin{quote}\fbox{\parbox{0.9\columnwidth}{
{\it If someone regretted that a particular thing happened} \\
{\it (A) that thing might or might not have happened} \\
{\it (B) that thing didn't happen} \\
{\it (C) that thing happened} {\bf [correct]}}}
\end{quote}
Predicting factuality requires understanding what the context
around an event implies about that event's occurrence. We
tested Aristo on this task using the veradicity question dev set
from DNC \cite{Poliak2018CollectingDN}, converted to multiple choice format (394 questions).
On this task, Aristo score 66.5\%, again suggesting Aristo has some
knowledge of how words affect the factiveness of the events that they modify.

\subsubsection{Counting}

Finally we ran Aristo on bAbI task 7, (a simple form of) counting \cite{babi},
converted to multiple choice with four options, as below. For example:
\begin{quote}\fbox{\parbox{0.9\columnwidth}{
{\it Daniel picked up the football. Daniel dropped the football. Daniel got the milk.}
{\it How many objects is Daniel holding? (A) zero (B) one (C) two (D) three}}}
\end{quote}
Aristo (again not fine-tuned on this dataset) did badly at this task,
scoring only 6\%.\footnote{Lower than random guessing, because Aristo frequently selects option D (``three''),
  an option which is (by chance) very rarely the right answer in this dataset.
  ``D'' is likely chosen due to a small random bias towards ``three'', and all questions
  looking stylistically similar. Note there is no training on this (nor other) probing datasets, so Aristo is
  unaware of the answer distribution.}
This results is perhaps not surprising, as this type of reasoning
is not exemplified in any way in an any of Aristo's training data.

\subsubsection{Scorecard}

\begin{figure}
\begin{center}
{\includegraphics[width=0.7\columnwidth]{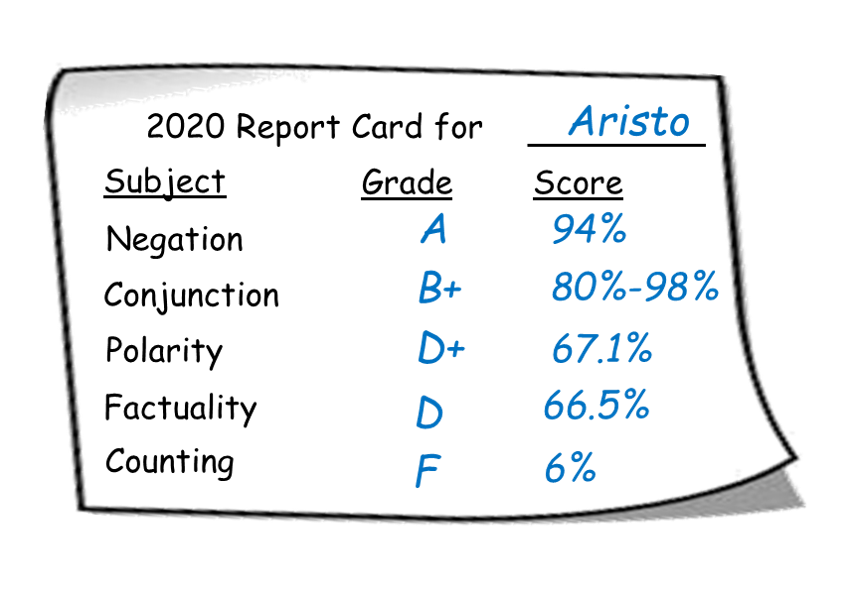}}
\end{center}
\vspace{-4mm}
\caption{Aristo, with no fine-tuning, passes probes for all but counting. \label{scorecard}}
\end{figure}

We can informally map these scores to a grade level to give Aristo
a Score Card (Figure~\ref{scorecard}). The most striking observation
is that Aristo ``passes'' all but counting, and has apparently acquired these skills
through its general fine-tuning on RACE and Science Exams, with no
fine-tuning at all on these specific probing datasets. Aristo
does appear to be doing more than just sentence matching, but not quite in
the systematic way a person would. These acquired latent skills are reflected
in the high scores Aristo achieves on the Science Exams.

\section{Discussion}

What can we conclude from this? Most significantly, Aristo has achieved surprising success on a formidable problem,
in particular by leveraging large-scale language models. The system thus serves as a demonstration of the
stunning progress that NLP technology as a whole has made in the last two years.


At the same time, exams themselves are an imperfect test of understanding science, and,
despite their many useful properties, are also only a partial test of machine
intelligence \cite{Davis2014TheLO}. Earlier, we highlighted several classes
of problems Aristo does not handle well, even though its exam scores are high: questions requiring diverse pieces
of evidence to be combined, reading comprehension (``story'') questions, meta-questions, and arithmetic.
Davis (\citeyear{Davis2016HowTW}) has similarly pointed out that as standardized tests are authored for people,
not machines, they also {\it don't} directly test for things that are easy for people
such as temporal reasoning, simple counting, and obviously impossible situations.
These are problem types that Aristo is not familiar with and would be hard for it (but not for people).
Clearly, science exams are just one of many different, partial indicators of progress in broader AI.
Finally, we have only been using multiple-choice questions, a format that
just requires ranking of answer choices, arguably allowing more use of weak evidence
compared with (say) generating an answer, or even independently deciding if an answer is true or false \cite{Clark2019BoolQET}.


On the other hand, we {\it do} see clear evidence of systematic
semantic skill in Aristo. For example, Aristo not only answers this question correctly:
\begin{quote}\fbox{\parbox{0.9\columnwidth}{
      {\it City administrators can encourage energy conservation by} \\
{\it (A) lowering parking fees} \\
{\it (B) decreasing the cost of gasoline.} \\
{\it (C) lowering the cost of bus and subway fares.} {\bf [correct,selected]}}}
\end{quote}
but flipping ``decreasing'' and ``lowering'' causes it to correctly change its answer:
\begin{quote}\fbox{\parbox{0.9\columnwidth}{
{\it City administrators can encourage energy conservation by} \\
{\it (A) lowering parking fees} \\
{\it (B) \sout{decreasing} {\bf increasing} the cost of gasoline.} {\bf [correct,selected]} \\
{\it (C) \sout{lowering} {\bf raising} the cost of bus and subway fares.}}}
\end{quote}
Our probes showed that such behavior is
not just anecdotal but systematic, suggesting that {\it some} form of reasoning
is occurring, but not in the traditional style of discrete symbol manipulation
in a formally designed language \cite{brachman1985readings,genesereth2012logical}.
Other work has similarly found that neural systems can learn systematic behavior \cite{scan,clark2020transformers}, and
these emergent semantic skills are a key contributor to Aristo's scores reaching the 90\% range.
Large-scale language model architectures have brought a dramatic, new
capability to the table that goes significantly beyond just pattern
matching and similarity assessment.

\section{Summary and Conclusion}

\begin{table}[t!]

\begin{center}
 \fbox{%
   \parbox{1\columnwidth}{
    \vspace{1mm}

\centerline{\underline{\bf Related Work on Standardized Testing for AI}}

\vspace{2mm}

\noindent
\underline{\bf Standardized Tests}

Standardized tests have long been proposed as challenge problems for AI \cite[e.g.,][]{bringsjord2003artificial,brachman2005selected,clark2016my,piatetsky2006grand},
as they appear to require significant advances in AI technology while also being accessible, measurable, understandable, and motivating. 

Earlier work on standardized tests focused on specialized tasks, for example, SAT word analogies \cite{turney2006similarity}, GRE word antonyms \cite{mohammad2013computing}, 
and TOEFL synonyms \cite{landauer1997solution}. More recently, there have been attempts at building systems to pass university entrance exams. Under NII's Todai project, several systems were developed for parts of the University of Tokyo Entrance Exam, including maths, physics, English, and history \cite{strickland2013can,nii-today,fujita2014overview}, although in some cases questions were modified or annotated before being given to the systems \cite[e.g.,][]{matsuzaki2014most}. Similarly, a smaller project worked on passing the Gaokao (China's college entrance exam) \cite[e.g.,][]{cheng2016taking,guo2017effective}. The Todai project was reported as ended in 2016, in part because of the challenges of building a machine that could ``grasp meaning in a broad spectrum'' \cite{todai-gives-up}.

\vspace{2mm}

\noindent
\underline{\bf Math Word Problems}

Substantial progress has been achieved on math word problems.  On plane geometry questions, \cite{Seo2015SolvingGP} demonstrated an approach that achieve a 61\% accuracy on SAT practice questions.  
The Euclid system \cite{Hopkins2017Euclid} achieved a 43\% recall and 91\% precision on SAT "closed-vocabulary" algebra questions, a limited subset of questions that nonetheless constitutes approximately 45\% of a typical math SAT exam.  Closed-vocabulary questions are those
 that do not reference real-world situations (e.g., "what is the largest prime smaller than 100?" or "Twice the product of x and y is 8. What is the square of x times y?")
 
 Work on open-world math questions has continued, but results on standardized 
 tests have not been reported and thus it is difficult to benchmark the progress relative to human performance.  See \namecite{Amini2019MathQATI} for a recent snapshot of the state of the art, and references to the literature on this problem.
}}
 \end{center}

  \end{table}

Answering science questions is a long-standing AI grand challenge \cite{reddy1988foundations,friedland2004project}.
We have described Aristo, the first system to achieve a score of over 90\% on the non-diagram, multiple choice
part of the New York Regents 8th Grade Science Exam, demonstrating that modern NLP methods can result in mastery of this task.
Although Aristo only answers multiple choice questions without diagrams,
and operates only in the domain of science,
it nevertheless represents an important milestone towards systems that can read
and understand. The momentum on this task has been remarkable, with accuracy moving from roughly 60\% to over 90\% in just three years.
In addition, the use of independently authored questions from a standardized test allows us to benchmark AI performance relative to human students.

Beyond the use of a broad vocabulary and scientific concepts, many of the benchmark questions intuitively appear to
require some degree of reasoning to answer. For many years in AI, reasoning was thought of as discrete
symbol manipulation. 
With the advent of deep learning, this notion of reasoning has
expanded,	
with systems performing challenging tasks using
neural architectures rather than explicit representation languages. Similarly, 
we observe surprising performance on answering science questions, and on specific semantic phenomena such as those probed earlier.
This suggests that the machine has indeed learned something about language and the world, and
how to manipulate that knowledge, albeit neither symbolically nor discretely.

Although an important milestone, this work is only a step on the long road toward a machine that
has a deep understanding of science and achieves Paul Allen's original dream of a Digital Aristotle.
A machine that has fully understood a textbook should not only be able to
answer the multiple choice questions at the end of the chapter - it should
also be able to generate both short and long answers to direct questions; it should be
able to perform constructive tasks, e.g., designing an experiment for a particular hypothesis;
it should be able to explain its answers in natural language and discuss them with a user;
and it should be able to learn directly from an expert who can identify and correct
the machine's misunderstandings. These are all ambitious tasks still largely beyond the
current technology, but with the rapid progress happening in NLP and AI, solutions may
arrive sooner than we expect.

\subsection*{Acknowledgements}
We gratefully acknowledge the late Paul Allen's inspiration, passion, and support for research on this grand challenge.
We also thank the many other contributors to Aristo, including
Niranjan Balasubramanian, Matt Gardner, Peter Jansen, Jayant Krishnamurthy,
Souvik Kundu, Todor Mihaylov, Harsh Trivedi, Peter Turney,
and the Beaker team at AI2, and to Ernie Davis, Gary Marcus, Raj Reddy,
and many others for helpful feedback on this work. We also thank
the anonymous reviewers for helpful comments that improved the paper.

\bibliography{references}
\bibliographystyle{aaai}

\end{document}